\documentclass[conference]{IEEEtran}
\usepackage[utf8]{inputenc}
\usepackage{xcolor}
\usepackage{csquotes}
\MakeOuterQuote{"}
\usepackage{amsmath}
\usepackage{url}
\usepackage{cite}
\usepackage{graphicx}
\usepackage{enumerate}
\usepackage{framed}
\usepackage{multirow}

\begin{document}
\title{MilkQA:\@ a Dataset of Consumer Questions\\for the Task of Answer Selection}

\author{
    \IEEEauthorblockN{Marcelo Criscuolo,
        Erick Rocha Fonseca\\ and Sandra Maria Aluísio}
\IEEEauthorblockA{
        University of São Paulo (USP)\\
        Institute of Mathematics and Computer Sciences\\
        São Carlos, Brazil\\
    \{mcrisc, erickrf, sandra\}@icmc.usp.br}
\and
\IEEEauthorblockN{Ana Carolina Sperança-Criscuolo}
\IEEEauthorblockA{São Paulo State University (Unesp)\\
        College of Letters and Sciences\\
        Araraquara, Brazil\\
        carolinasperanca@fclar.unesp.br}
}


\maketitle

\begin{abstract}
        We introduce MilkQA, a question answering dataset from the dairy domain dedicated to the study of consumer questions. The dataset contains 2,657 pairs of questions and answers, written in the Portuguese language and originally collected by the Brazilian Agricultural Research Corporation (Embrapa). All questions were motivated by real situations and written by thousands of authors with very different backgrounds and levels of literacy, while answers were elaborated by specialists from Embrapa's customer service. Our dataset was filtered and anonymized by three human annotators. Consumer questions are a challenging kind of question that is usually employed as a form of seeking information. Although several question answering datasets are available, most of such resources are not suitable for research on answer selection models for consumer questions. We aim to fill this gap by making MilkQA publicly available. We study the behavior of four answer selection models on MilkQA:\@ two baseline models and two convolutional neural network archictetures.  Our results show that MilkQA poses real challenges to computational models, particularly due to linguistic characteristics of its questions and to their unusually longer lengths.  Only one of the experimented models gives reasonable results, at the cost of high computational requirements. 
\end{abstract}

%
\IEEEpeerreviewmaketitle

\section{Introduction}
In question answering, the task of answer selection consists in finding the best answer for a question in a pool of pre-selected candidate answers. More formally, given a question $q$ and a set of candidate answers $C_q = \{a_1, a_2, \ldots, a_n\}$, the goal is to find a candidate answer $a_i$ that belongs to $G_q$, where $G_q \subset C_q$ is the ground truth set of question $q$.

Recent research on answer selection models has been focused on objective and well-formed factoids, while little attention has been given to other kinds of questions, such as consumer questions~\cite{zhang2010,kilicoglu2013ellipsis}, which are a common form of seeking information that usually occurs in Q\&A community sites, forums and customer services.

Consumer questions differ from factoids mainly in their structure.  They are often formed by multiple sentences, that are related to each other at semantic level by textual cohesion mechanisms.  A striking feature of consumer questions is the presence of a problem description (context) that is followed by one or more sub-questions.  Those sub-questions are short, and seem incomplete when viewed in isolation, since they rely strongly on information provided previously in the context.  Occasionally, the context is not given explicitly, and the consumer question is posed as a sequence of interrelated, complementary sub-questions.  Another noticeable characteristic of this kind of question is that sub-questions are often posed as declarative sentences beginning with \emph{"I would like to know"}~\cite{zhang2010}, instead of direct interrogative sentences.

We define a consumer question as a text segment that fulfill one of the following criteria:
(a) contains a context, or problem description, and sub-questions that rely on that previous information to be completely understood;
(b) contains two or more related sub-questions that reinforce each other.
By this definition, the context description is not mandatory and what really characterizes a text segment as a consumer question are the relations of complementary meaning between sentences.

Consumer questions pose several challenges to computational models.  They often contain misspelings, poor punctuation, and layman's terms that do not match the vocabulary of potential answers~\cite{zhang2010,kilicoglu2013ellipsis}.  Context descriptions commonly contain details in excess, which make it hard to distinguish information that is really relevant~\cite{lei2016gated-convolutions}.  Finally, since consumer questions are usually much longer than factoid questions, answer selection models tend to require a great deal of computational resources.

Several datasets are available for the development of answer selection models~\cite{wang2007qasent,yang2015wikiqa,feng2015insuranceqa,squad2016,selqa2016,nguyen2016msmarco}.  However, those resources are not suitable for the development of answer selection models for consumer questions, since they are usually focused on short interrogative sentences.  Moreover, questions in some datasets are elaborated artificially and do not represent the real use of natural language.  Annotators are asked to elaborate questions for given pieces of text, for example.  The result of such a process are simple questions of limited practical use.

Realistic datasets have been responsible for driving fields forward.  The contribution of Penn Treebank~\cite{marcus1993penn} for the field of syntatic parsing is a known example.  In this paper, we present MilkQA\footnote{MilkQA is publicly available at: \url{http://nilc.icmc.usp.br/nilc/index.php/milkqa}}, a realistic dataset of consumer questions from the dairy domain in the Portuguese language.  This dataset was formed from thousands of consumer questions collected within the period of twelve years by the customer service of an important agricultural research institution.  Every question comes from real situations and is answered in detail by a specialist.  The current version of MilkQA contains 2,657 pairs of messages (questions and respective answers) selected and anonymized by three annotators.  The average length of questions is 57 words, while the average length of questions in factoid datasets, like WikiQA, is less than 10 words (see Table~\ref{tab:question-lengths}).  Answers in MilkQA are even longer than questions, and may have hundreds of words.

We implemented four answer selection models to study their behavior in MilkQA\@. The performance of two state-of-the-art Convolutional Neural Network (CNN) models is compared to results achieved by two baseline models. The first baseline is idf-weighted word matching, and the second is the cosine of idf-weighted sums of word embeddings. We experiment with MilkQA and with the short-question dataset WikiQA~\cite{yang2015wikiqa}.  Although relatively simple models achive good results on the short-question dataset, their results on MilkQA are only modest. Better results are achieved by a complex model, though it imposes high requirements on computational resources.

This paper is organized as follows.  In Section~\ref{sec:relatedwork}, we review the main datasets that are available for the task of answer selection.  In Section~\ref{sec:milkqa}, we describe in detail our new dataset, MilkQA\@. In Section~\ref{sec:experiments}, we present and discuss our experiments.  Finally, the Section~\ref{sec:conclusion} is dedicated to the conclusion and future work.

\section{Related Work}
\label{sec:relatedwork}
Although some large question answering datasets are available (Table~\ref{tab:datasets}), most are composed of rather simple questions, either formulated artificially or derived from queries submitted to search engines.  SelQA~\cite{selqa2016} was introduced as a benchmark dataset for the task of answer selection.  It contains nearly 8 thousand questions, formulated by crowdsourcing workers from text segments extracted from Wikipedia.  The same process was used to build SQuAD~\cite{squad2016}, which is much larger, with 100 thousand questions.  Questions derived from previously known text segments are generally simpler than natural questions and share many words with their answers.  One significant concern with this approach is that the lexical overlap will make sentence selection easier and might inflate the performance of systems~\cite{yang2015wikiqa}.

Collecting queries submitted to search engines is an alternative to avoid artificial questions.  WikiQA~\cite{yang2015wikiqa} is a dataset of natural factoid questions submitted by users of the Bing search engine.  Each question is related to a set of candidate answer sentences extracted from Wikipedia articles, where correct answers are identified by human annotators.  Part of the questions in this dataset does not have correct answers in their candidate sets.  These examples are useful for the task of answer triggering, where models are required to identify the lack of appropriate answers for a question in its candidate set.  The process of collecting questions from search engine logs was also used to build MS MARCO~\cite{nguyen2016msmarco}, which may be regarded as a larger version of WikiQA\@. MS MARCO contains 100 thousand factoid questions, as well as questions with no correct answers.  However, differently from WikiQA, answers for MS MARCO questions were not restricted to Wikipedia, but were collected from thousands of web documents.

Another important dataset is TREC-QA~\cite{wang2007qasent}, also known as QASent, which became the standard benchmark dataset for the answer selection task.  It contains only 227 questions, chosen from the Text REtrieval Conference (TREC) QA dataset.  To generate candidate answer sets, the authors selected sentences from each question's document pool that contained one or more non-stopwords from the question.  Thus, TREC-QA also show significant lexical overlap between questions and answers, inducing a strong bias on models based on word matching.

Finally, InsuranceQA~\cite{feng2015insuranceqa} is a non-factoid dataset from the insurance domain.  It was created from data collected from an Internet site, where insurance experts answer questions received from users.  While answers in InsuranceQA are detailed explanations, significantly longer than typical answers in other datasets, the source site limits questions to single sentences, making them short and objective, with few words only (see Tables~\ref{tab:question-lengths} and~\ref{tab:answer-lengths}).

Question answering datasets in the Portuguese language are less abundant.  The Págico~\cite{simoes2012pagico} dataset was created for a shared task on information retrieval, organized for the Portuguese language.  It contains only 153 manually formulated factoid questions, that are not particularly adequate for question-answer matching techniques, typically used in the answer selection task.

\begin{table}[!t]
\caption{Answer Selection Datasets}
\label{tab:datasets}
\centering
        \begin{tabular}{l | l | r }
                \hline
                \textbf{Dataset} & \textbf{Question Source} & \textbf{\# Questions}\\
                \hline
                SelQA & crowdsourcing & 7,904\\
                SQuAD & crowdsourcing & 100K\\
                WikiQA & user query logs & 3,047\\
                MS MARCO & user query logs & 100K\\
                TREC-QA & user query logs + editor & 227\\
                InsuranceQA & users (single sentences) & 16,889\\
                Págico & editors & 153\\
                \hline
                MilkQA & consumer emails & 2,657\\
                \hline
        \end{tabular}
\end{table}

\begin{figure}[!t]
\centering
\includegraphics[width=0.48\textwidth]{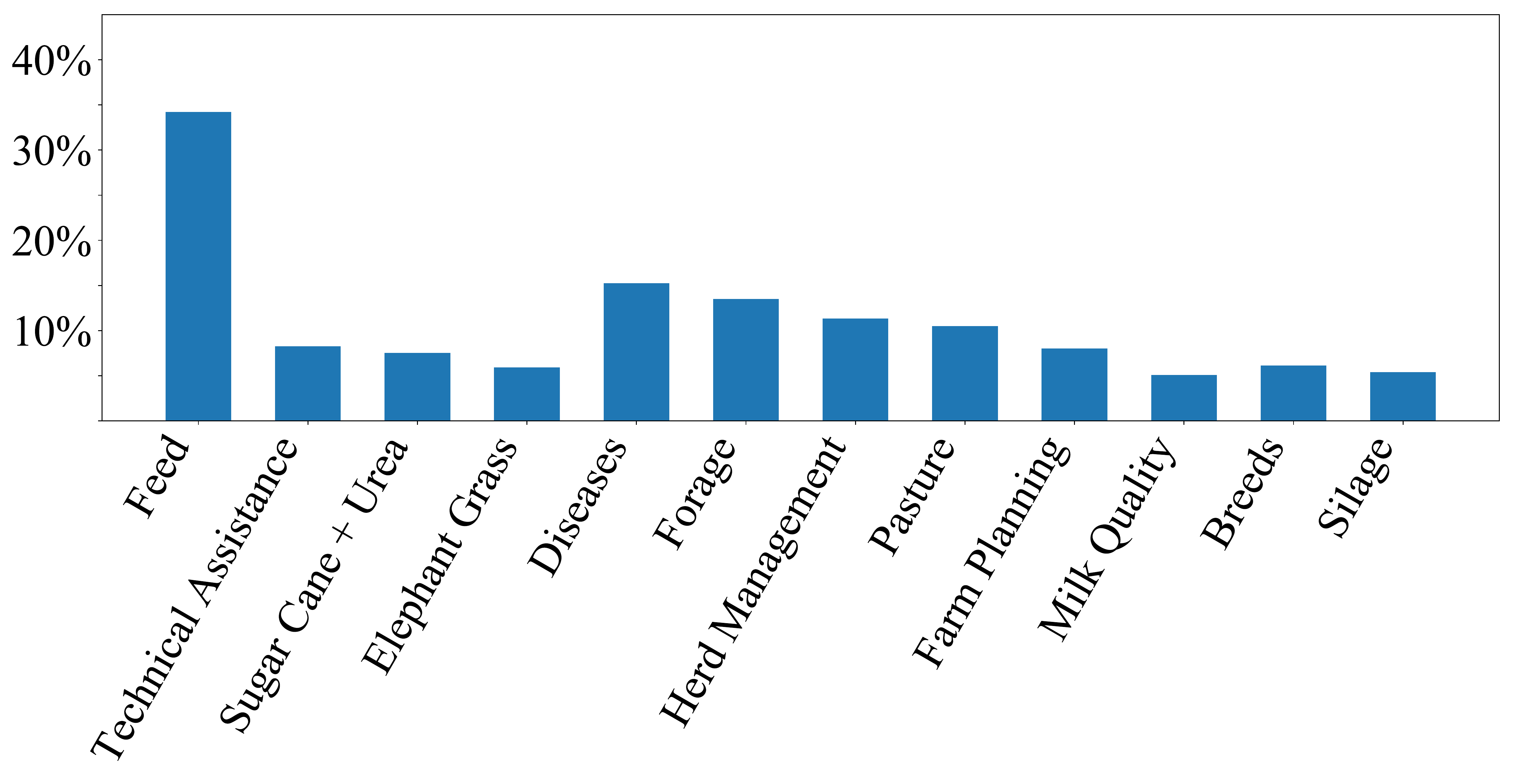}
\caption{Label frequencies in the MilkQA dataset. Labels with frequencies below 5\% are not shown.}
\label{fig:label-freq}
\end{figure}

\section{MilkQA Dataset}
\label{sec:milkqa}
In this section, we describe the process of creating our dataset, MilkQA, comparing its unique features to those of other datasets.

\subsection{Data Collection and Preparation}
MilkQA data was originally collected by the Brazilian Agricultural Research Corporation (Embrapa).  As all of Embrapa's research unities, Embrapa Dairy Cattle maintains a customer service dedicated to assist any citizen interested in their business.  This service gets many email messages, that are archived with their respective responses, after getting labels that describe their contents, such as \texttt{Breed} and \texttt{Diseases}. There are 58 different labels. On average, two labels are applied to each message\footnote{MilkQA may also be an interesting dataset for the problem of multi-label classification~\cite{tsoumakas2010mlc}.}.

Our dataset is derived from a message archive created by Embrapa Dairy Cattle's customer service, containing nearly 27 thousand message pairs (requests and responses), collected from the year 2003 to 2012.  Questions are written by thousands of authors with very different backgrounds and levels of literacy, while answers are elaborated by customer services specialists.

The customer service works as a counter that gets all sorts of requests, ranging from questions about dairy cattle to job applications.  However, to build our answer selection dataset, only messages containing consumer questions were considered.  Thus, many messages in the archive, that were not knowledge requests, had to be discarded.  The filtering process was carried out in two phases.  First, we used an automated process to discard most noisy and unwanted messages.  Then, the remaining messages went through careful manual selection and cleaning.

The automated process relied on the labels to decide which messages to discard.  We estimated usage rates for each label, by manually analyzing a random sample of messages drawn from the full archive.  This statistical analysis showed that messages marked with some labels -- such as \texttt{Training} and \texttt{Internship}, for example -- were rarely used.  Such messages were automatically discarded. A number of duplicated messages were identified and removed as well.  The resulting pre-selected archive contained about 10 thousand messages, that needed to go through manual selection.

In the second phase, three annotators worked on the selection of remaining messages, performing two activities simultaneously.  At the same time they rejected non-consumer questions, they also anonymized and cleaned accepted messages. Message cleaning consisted in removing particular data, such as people names and contact information, and data that was not related to questions, such as corporate signatures and advertisements, typically found in email messages.

\begin{figure}[!t]
        \begin{framed}
                \begin{enumerate}[(a)]
                        \item What causes destruction of the ozone layer?
                        \item What is the mortality rate for lightning strikes?
                        \item What are the different types of homeowners insurance?
                \end{enumerate}
        \end{framed}
\caption{Examples of objective questions drawn from datasets WikiQA (a, b) and InsuranceQA (c).}
\label{fig:objective-questions}
\end{figure}

\begin{figure}[!t]
        \begin{framed}  
                I have some dairy cows and noticed that the milk production have dropped a lot.
                I took tests and noticed the presence of mastitis.
                (a) \emph{I do not know which medication to apply.}
                (b) \emph{What is the best treatment?}
                (c) \emph{What is the quickiest way to get production back to normal?}
        \end{framed}
        \caption{A typical consumer question from MilkQA\@. This question was originally written in Portuguese.}
\label{fig:milkqa-example}
\end{figure}

\begin{figure}[!t]
        \begin{framed}
                \begin{enumerate}[(a)]
                        \item Can calves eat oat? Is it better to mix some mineral?
                        \item Can I apply ivermectin to lactating cows? Is it bad for the calf? What about humans?
                \end{enumerate}
        \end{framed}
\caption{Consumer questions with no explicit context.}
\label{fig:nocontext-questions}
\end{figure}

For this first version of MilkQA, about half of the messages in the pre-selected archive were examined by the annotators. Approximately 53\% of the examined messages were selected for the dataset, which means MilkQA currently contains 2,657 anonymized message pairs.  We computed the frequency of labels applied to these messages and showed the most frequent on the graph in \figurename~\ref{fig:label-freq}.  The graph shows, for example, that the label \texttt{Feed} was applied to almost 35\% of the messages.

Those labels provide an overview of common message contents.  The frequency of \texttt{Elephant Grass}, \texttt{Forage} and similar labels reveal that there is a high number of questions related to herd feeding, for example.  Other labels, like \texttt{Farm Planning} and \texttt{Herd Management}, show that milk production is also a major concern.  Indeed, label frequencies clearly reflect that questions were motivated by the interest of consumers in obtaining knowledge about the dairy cattle domain.

\subsection{Consumer Questions in MilkQA}
To highlight the features of MilkQA questions, we contrast examples of consumer questions extracted from this dataset with common examples of objective questions from other datasets.  The examples shown in \figurename~\ref{fig:objective-questions} were extracted from WikiQA and InsuranceQA, and illustrate common features of objective questions.  They are represented by single interrogative sentences that are short, direct and complete.  In \figurename~\ref{fig:milkqa-example}, we show a question drawn from MilkQA\@. As a typical consumer question, it is represented as a short text containing multiple sentences.

\begin{table}[!t]
        \caption{Comparison of Question Length Statistics}
        \label{tab:question-lengths}
        \centering
        \begin{tabular}{l | r | r | r | r | r}
                \hline
                \multirow{2}{*}{\bfseries Dataset} &
                \multicolumn{5}{c}{\bfseries Words per Question} \\
                \cline{2-6}
                 & min & avg & 50-p & 99-p & max \\
                \hline
                WikiQA & 2 & 7 & 7 & 16 & 23 \\
                InsuranceQA & 2 & 7 & 7 & 14 & 57 \\
                MilkQA & 4 & 57 & 46 & 217 & 681 \\
                \hline
        \end{tabular}
\end{table}

\begin{table}[!t]
        \caption{Comparison of Answer Length Statistics}
        \label{tab:answer-lengths}
        \centering
        \begin{tabular}{l | r | r | r | r | r}
                \hline
                \multirow{2}{*}{\bfseries Dataset} &
                \multicolumn{5}{c}{\bfseries Words per Answer} \\
                \cline{2-6}
                 & min & avg & 50-p & 99-p & max \\
                \hline
                WikiQA & 1 & 22 & 21 & 57 & 165 \\  
                InsuranceQA & 15 & 100 & 78 & 395 & 1,180 \\
                MilkQA & 8 & 237 & 157 & 688 & 3,427 \\
                \hline
        \end{tabular}
\end{table}

The average length of consumer questions is usually several times longer than that of objective questions.  In Table~\ref{tab:question-lengths}, we present statistics about question lengths in MilkQA compared to statistics of other two answer selection datasets.  Mininum, average and maximum question lengths are shown for each dataset.  The columns 50-p and 99-p refer to the length of questions in the 50th and 99th percentile, respectively.  The table shows, for instance, that 99\% of the questions in WikiQA contains 16 words or less.  MilkQA contains a few non-consumer questions that are really short. Those are responsible for the small number of words shown in the first column of the table.  On the other hand, MilkQA contains some really long questions, with hundreds of words.  Sub-questions are found among full problem descriptions in such cases.  Table~\ref{tab:answer-lengths} show analogous statistics for MilkQA answers, which are even longer than questions.

Consumer questions are usually composed of a context description and one or more sub-questions. Most of the questions in MilkQA follow this pattern. The example in \figurename~\ref{fig:milkqa-example} contains three sub-questions focused on the problem of decrease in milk production, caused by an specific disease (mastitis). Furthermore, many sub-questions in MilkQA are expressed indirectly, like the example sentence in \figurename~\ref{fig:milkqa-example}(a), which implies the question \emph{"Which medication should I apply?"}. Another indirect construction that is very frequent in the dataset is \emph{"I would like to know\dots"}. MilkQA also contains consumer questions that provide no explicit context, like the two examples in \figurename~\ref{fig:nocontext-questions}, composed of several interrelated interrogative questions.

It is clear from the examples in Figures~\ref{fig:milkqa-example} and~\ref{fig:nocontext-questions} that sub-questions alone rarely have complete meanings. The meaning of one sub-question -- like \figurename~\ref{fig:milkqa-example}(b), for instance -- usually depends on information given in the context, or found in other sub-questions. This dependency occurs by means of linguistic mechanisms, such as substitution and ellipsis~\cite{halliday1976}, which are really challenging for computational processing.

Other common observations in questions from MilkQA are misspellings, typos and poor use of grammar and punctuation.  All of these may have strong impact over tools such as syntactic parsers. Answers in turn are generally well written texts and show few writting problems. In fact, most answers are detailed technical explanations that tend to be reused with small changes made to better meet particular question needs.

\subsection{Answer Selection Pool}
The task of answer selection requires a pool of candidate answers associated with every question in the dataset. For each question, one or more ground truth answers must be included in the pool.

To build candidate pools from the (question, answer) pairs in MilkQA, we performed a cluster analysis on the full answer set. We identified some clusters of almost identical answers, with only minor differences between elements. This fact is due to the use of answer templates in the customer service. Such nearly identical answers can cause problems to answer selection models if a ground truth answer conflicts with a negative candidate in the same pool.  To avoid this problem, we do not allow more than one answer from the same cluster in each pool. We used a density-based algorithm with parameters tuned to generate very tight clusters. Answers are represented by tf-idf vectors, that are reduced by an autoencoder to 100 dimensions. This analysis identified 97 clusters, containing an average of 6 answers each. The largest and the smallest clusters contains 104 and 2 answers, respectively. In fact, 75\% of the identified clusters contains 4 answers or less. That means large clusters represent 25\% of the total.

Each answer pool contains 50 candidates, including one ground truth answer, which was originally provided by the customer service for the corresponding question. The other candidates are the answers nearest to the ground truth, starting from an initial distance determined empirically. Such a strategy aims to reproduce challenging situations where answer selection models have to distinguish between similar answers.

\section{Experiments}
\label{sec:experiments}
We evaluated the behavior of baseline and state-of-the-art answer selection models on WikiQA and MilkQA\@. All models approach answer selection as a ranking task, where each candidate answer is assigned a relevance score. The highest score should indicate the correct answer. Standard metrics Precision at top one\footnote{This is precision at $k$, with $k = 1$.} (P@1) and Mean Average Precision (MAP) are adopted for performance evaluation.

\subsection{Answer Selection Models}
We consider two baseline models: Weighted Word Matching (WWM) and Weighted Sum of word Embeddings (WSE). The first model computes the sum of IDF values for each non-stopword in the question that also occurs in the answer, while the second computes the cosine similarity between question and answer represented as IDF-weighted sums of their word vectors.

We also implemented two Convolutional Neural Network (CNN) models. The first, CNN-STD, is a simplified version of the ranking model proposed in~\cite{severyn-moschitti2015}. The input to this model are two matrices representing a question and a candidate answer. Each matrix is initialized with pre-trained word embeddings and mapped to a feature vector by a convolutional layer. The two feature vectors are concatenated and passed to a hidden layer. A relevance score is obtained by applying the sigmoid function to the hidden layer output.  Our implementation differs from the original model in that it does not consider additional features, neither an intermediate similarity matrix.  We use hyperbolic tangent for non-linearity and dropout is applied to the fully-connected layer for regularization.

The second model, CNN-LDC, is implemented exactly as described in~\cite{wang2016ldc}. The general architecture resembles the previous model, but CNN-LDC employs two-channel convolutions. Before mapping matrices to feature vectors, the model decomposes each word in two components that capture semantic similarities and dissimilarities between the current question and candidate answer. This lexical decomposition relies on an attention matrix computed for each input pair.

In all experiments, we used 300-dimensional word vectors trained with word2vec~\cite{mikolov2013-distributed}, using the Continous Bag-of-Words model (CBOW)~\cite{mikolov2013-efficient}. For WikiQA (English), we used the freely available vectors from Google News\footnote{Available at \url{https://code.google.com/p/word2vec/}}, and for MilkQA (Portuguese), we used vectors trained by our research group\footnote{Available at \url{http://nilc.icmc.usp.br/embeddings}} in a wide range of sources, like the LX-Corpus~\cite{rodriguesetal2016}, texts crawled from the Portuguese versions of Wikipedia and Google News, movie subtitles, newspaper articles, and children's story books. All the text sources used to train Portuguese word vectors total approximately 1.4 billion tokens.

Hyperparameters are the same for both CNN models. Filter lengths are 1, 2, and 3, with 50 feature maps each, dropout uses a drop-rate of 0.20, and Adam optimizer is used to minimize the squared errors. In the training sets, answers are labeled with 1 if they are correct or 0 otherwise. Training is interrupted by early stopping if no performance improvement is observed after two evaluations of the development set. Batches are single triples of form $(q_i, a^+, a^-)$, where $a^+$ is a ground truth answer for question $q_i$, and $a^-$ is a random incorrect answer selected from the question pool. To build another triple (batch), the training algorithm selects the next question, $q_{i+1}$. That means a given question is used only once in each epoch.  The maximum length for questions and answers are, respectively, 20 and 40 for WikiQA, and 315 and 710 for MilkQA\@. 

The dataset was partitioned into train, dev and test subsets containing 2,307, 50 and 300 questions, respectively.  The choice for the test set size aims to keep most examples on the the training set, while the dev set size was chosen to avoid slowing down the training process.  We found 50 examples to be a good compromise between performance assessment and computation time.

\begin{table}[!t]
        \caption{Experiment Results}
        \label{tab:results}
        \centering
        \begin{tabular}{l | r r | r r}
                \hline
                \multirow{2}{*}{\bfseries Model} &
                \multicolumn{2}{c|}{\bfseries WikiQA} & \multicolumn{2}{c}{\bfseries MilkQA} \\
                \cline{2-5}
                & P@1 & MAP & P@1 & MAP \\
                \hline
                WWM & 0.5062 & 0.5100 &  0.2467 & 0.3836 \\
                WSE & 0.3951 & 0.5838 & 0.1733 &  0.2552 \\
                CNN-STD & 0.4135 & 0.5746 & 0.4100 & 0.5573 \\
                CNN-LDC & 0.5485 & 0.6848 & 0.5700 & 0.6899 \\
                \hline
        \end{tabular}
\end{table}

\subsection{Results}
Table~\ref{tab:results} summarizes the results of our experiments. Performance is measured with P@1, which is computed by our own evaluation script, and with MAP, computed by the official TREC scorer\footnote{We used version 8.1, available at \url{http://trec.nist.gov/trec_eval/}} (\texttt{trec\_eval}). These two metrics reflect different system capabilities. MAP measures how good is a system at placing correct answers at top rank positions, while P@1 represents the fraction of questions that are correctly answered, with a ground truth answer assigned exactly to the first rank position.

As shown in the first row of Table~\ref{tab:results}, the word matching baseline (WWM) achieves good results on the WikiQA dataset and outperforms CNN-STD, according to P@1. This good performance may be due to frequent lexical overlap in WikiQA\@. Word matching shows much lower performance on MilkQA\@.

The best performance on both datasets is achieved by \mbox{CNN-LDC}, while WSE gives the worst overall results. The very low scores of WSE on MilkQA suggests that the weighted sum of vectors is not powerful enough to capture the semantics of longer text segments, such as MilkQA questions and answers. Although word matching (WWM) performs better than WSE on MilkQA, both baselines give very low results compared to CNN models. In fact, \mbox{CNN-LDC} significantly outperforms the other models on this dataset.

Despite the good MAP scores achieved by CNN-LDC on both datasets, compared to other models, P@1 still indicates large room for improvement (see results of the literature on answer selection models~\cite{wang2007qasent,yang2015wikiqa,feng2015insuranceqa,squad2016,selqa2016,nguyen2016msmarco}). Even on MilkQA, where P@1 score was higher, the value 0.57 may be interpreted as only 57\% of the questions being correctly answered by the best of the models.

\subsection{Discussion}
At first glance, the lower number of training samples in MilkQA, compared to those of other datasets (Table~\ref{tab:datasets}), may seem to be the cause for achieving modest results in the experiments (Table~\ref{tab:results}). However, much higher results have been achieved on very small datasets. CNN-LDC, for instance, achieves a MAP score of 0.77 in TREC-QA~\cite{wang2016ldc}, whose size is only a fraction of the size of MilkQA\@. Thus, we believe the unique features of consumer questions are the real cause of the modest results observed in the experiments.

MilkQA features also imposes severe restrictions to answer selection models.  The length of questions and answers limits the size of traning batches as well as the number and length of convolution filters in CNNs. To deal with this obstacle, very conservative parameters are chosen for our models. For instance, while CNN-LDC is trained with 500 feature maps in the original paper, we train our model only with 150 feature maps to reduce memory consumption and lower the number of model parameters that should be learned. We also truncate questions and answers to avoid the waste of resources caused by some very long outliers. However, we tried to keep the greatest possible number of texts untouched by choosing cutoff lengths that cause the truncation of only 0.3\% of the examples in the dataset. We also truncate sentences in WikiQA, so we can have results comparable to those reported on the CNN-LDC paper~\cite{wang2016ldc}.

Training and evaluation of CNN-LDC on MilkQA take long periods of time, even running the processes on GPUs. To compute scores for 50 candidate answers, this model takes about 20 seconds. At this speed, 1.67 hours were taken to evaluate the full test set, while 6.7 hours were necessary to train the model. To reduce training time, we limited the dev set size to only 50 questions, and evaluate the model each 500 batches to decide on early stopping. Each evaluation round of this tiny dev set takes around 16 minutes.

\section{Conclusion}
\label{sec:conclusion}
We introduced MilkQA, a dataset of consumer questions for the task of answer selection. MilkQA contains 2,657 pairs of real questions and answers in the Portuguese language, asked by a large number of authors of different levels of literacy. MilkQA pose difficult challenges to answer selection models due to the linguistic characteristics of its questions and to their much longer length, compared to traditional QA datasets. In our experiments, only modest results could be achieved by simple answer selection models on MilkQA, while a complex model could achieve better results at the cost of high consumption of computational resources. We hope that MilkQA will contribute to further develop research on answer selection involving consumer questions.

We plan to release a new version of MilkQA in the future, which we expect to contain twice the number of questions in the current version. To achieve this goal, we intend to continue the work of message cleaning and anonymization that is carried out by human annotators.

\section*{Acknowledgment}
The work of Marcelo Criscuolo was funded by Federal Institute of Education, Science and Technology of São Paulo (IFSP).
The work of Erick Fonseca was funded by Fapesp grant number 2013/22973-0.
The authors are grateful to Embrapa Dairy Cattle for providing the MilkQA data.

\bibliographystyle{IEEEtran}
\bibliography{references}

\begin{thebibliography}{10}
\providecommand{\url}[1]{#1}
\csname url@samestyle\endcsname
\providecommand{\newblock}{\relax}
\providecommand{\bibinfo}[2]{#2}
\providecommand{\BIBentrySTDinterwordspacing}{\spaceskip=0pt\relax}
\providecommand{\BIBentryALTinterwordstretchfactor}{4}
\providecommand{\BIBentryALTinterwordspacing}{\spaceskip=\fontdimen2\font plus
\BIBentryALTinterwordstretchfactor\fontdimen3\font minus
  \fontdimen4\font\relax}
\providecommand{\BIBforeignlanguage}[2]{{%
\expandafter\ifx\csname l@#1\endcsname\relax
\typeout{** WARNING: IEEEtran.bst: No hyphenation pattern has been}%
\typeout{** loaded for the language `#1'. Using the pattern for}%
\typeout{** the default language instead.}%
\else
\language=\csname l@#1\endcsname
\fi
#2}}
\providecommand{\BIBdecl}{\relax}
\BIBdecl

\bibitem{zhang2010}
\BIBentryALTinterwordspacing
Y.~Zhang, ``Contextualizing consumer health information searching: An analysis
  of questions in a social {Q}\&{A} community,'' in \emph{Proceedings of the
  1st ACM International Health Informatics Symposium}, ser. IHI '10.\hskip 1em
  plus 0.5em minus 0.4em\relax New York, NY, USA: ACM, 2010, pp. 210--219.
  [Online]. Available: \url{http://doi.acm.org/10.1145/1882992.1883023}
\BIBentrySTDinterwordspacing

\bibitem{kilicoglu2013ellipsis}
\BIBentryALTinterwordspacing
H.~Kilicoglu, M.~Fiszman, and D.~Demner-Fushman, ``Interpreting consumer health
  questions: The role of anaphora and ellipsis,'' in \emph{Proceedings of the
  2013 Workshop on Biomedical Natural Language Processing}.\hskip 1em plus
  0.5em minus 0.4em\relax Association for Computational Linguistics, 2013, pp.
  54--62. [Online]. Available: \url{http://aclweb.org/anthology/W13-1907}
\BIBentrySTDinterwordspacing

\bibitem{lei2016gated-convolutions}
\BIBentryALTinterwordspacing
T.~Lei, H.~Joshi, R.~Barzilay, T.~Jaakkola, K.~Tymoshenko, A.~Moschitti, and
  L.~M{\`a}rquez, ``Semi-supervised question retrieval with gated
  convolutions,'' in \emph{Proceedings of the 2016 Conference of the North
  American Chapter of the Association for Computational Linguistics: Human
  Language Technologies}.\hskip 1em plus 0.5em minus 0.4em\relax Association
  for Computational Linguistics, 2016, pp. 1279--1289. [Online]. Available:
  \url{http://aclweb.org/anthology/N16-1153}
\BIBentrySTDinterwordspacing

\bibitem{wang2007qasent}
\BIBentryALTinterwordspacing
M.~Wang, N.~A. Smith, and T.~Mitamura, ``What is the {J}eopardy model? a
  quasi-synchronous grammar for {QA},'' in \emph{Proceedings of the 2007 Joint
  Conference on Empirical Methods in Natural Language Processing and
  Computational Natural Language Learning (EMNLP-CoNLL)}.\hskip 1em plus 0.5em
  minus 0.4em\relax Prague, Czech Republic: Association for Computational
  Linguistics, June 2007, pp. 22--32. [Online]. Available:
  \url{http://www.aclweb.org/anthology/D/D07/D07-1003}
\BIBentrySTDinterwordspacing

\bibitem{yang2015wikiqa}
\BIBentryALTinterwordspacing
Y.~Yang, W.-t. Yih, and C.~Meek, ``{WikiQA}: A challenge dataset for
  open-domain question answering,'' in \emph{Proceedings of the 2015 Conference
  on Empirical Methods in Natural Language Processing}.\hskip 1em plus 0.5em
  minus 0.4em\relax Association for Computational Linguistics, 2015, pp.
  2013--2018. [Online]. Available: \url{http://aclweb.org/anthology/D15-1237}
\BIBentrySTDinterwordspacing

\bibitem{feng2015insuranceqa}
\BIBentryALTinterwordspacing
M.~Feng, B.~Xiang, M.~R. Glass, L.~Wang, and B.~Zhou, ``Applying deep learning
  to answer selection: A study and an open task,'' in \emph{2015 IEEE Workshop
  on Automatic Speech Recognition and Understanding (ASRU)}, Dec 2015, pp.
  813--820. [Online]. Available: \url{https://arxiv.org/abs/1508.01585}
\BIBentrySTDinterwordspacing

\bibitem{squad2016}
\BIBentryALTinterwordspacing
P.~Rajpurkar, J.~Zhang, K.~Lopyrev, and P.~Liang, ``{SQuAD}: 100,000+ questions
  for machine comprehension of text,'' in \emph{Proceedings of the 2016
  Conference on Empirical Methods in Natural Language Processing}.\hskip 1em
  plus 0.5em minus 0.4em\relax Association for Computational Linguistics, 2016,
  pp. 2383--2392. [Online]. Available:
  \url{http://aclweb.org/anthology/D16-1264}
\BIBentrySTDinterwordspacing

\bibitem{selqa2016}
\BIBentryALTinterwordspacing
T.~Jurczyk, M.~Zhai, and J.~D. Choi, ``{SelQA: A New Benchmark for
  Selection-based Question Answering},'' in \emph{Proceedings of the 28th
  International Conference on Tools with Artificial Intelligence}, ser.
  ICTAI'16, San Jose, CA, 2016. [Online]. Available:
  \url{https://arxiv.org/abs/1606.08513}
\BIBentrySTDinterwordspacing

\bibitem{nguyen2016msmarco}
\BIBentryALTinterwordspacing
T.~Nguyen, M.~Rosenberg, X.~Song, J.~Gao, S.~Tiwary, R.~Majumder, and L.~Deng,
  ``{MS MARCO}: A human generated machine reading comprehension dataset,'' in
  \emph{Proceedings of the 30th Conference on Neural Information Processing
  Systems (NIPS 2016)}, Barcelona, Spain, 2016. [Online]. Available:
  \url{https://arxiv.org/abs/1611.09268v2}
\BIBentrySTDinterwordspacing

\bibitem{marcus1993penn}
M.~P. Marcus, M.~A. Marcinkiewicz, and B.~Santorini, ``Building a large
  annotated corpus of english: The penn treebank,'' \emph{Computational
  Linguistics}, vol.~19, no.~2, pp. 313--330, 1993.

\bibitem{simoes2012pagico}
A.~Simões, L.~Costa, and C.~Mota, ``Tirando o chapéu à {Wikipédia}: A
  coleção do {Págico} e o {Cartola},'' \emph{Linguamática}, vol.~4, no.~1,
  pp. 19--30, 2012.

\bibitem{tsoumakas2010mlc}
G.~Tsoumakas, I.~Katakis, and I.~Vlahavas, ``Mining multi-label data,''
  \emph{Data mining and knowledge discovery handbook}, pp. 667--685, 2010.

\bibitem{halliday1976}
M.~A.~K. Halliday and R.~Hasan, \emph{Cohesion in {English}}.\hskip 1em plus
  0.5em minus 0.4em\relax Longman, 1976.

\bibitem{severyn-moschitti2015}
A.~Severyn and A.~Moschitti, ``Learning to rank short text pairs with
  convolutional deep neural networks,'' in \emph{Proceedings of the 38th
  International ACM SIGIR Conference on Research and Development in Information
  Retrieval}.\hskip 1em plus 0.5em minus 0.4em\relax ACM, 2015, pp. 373--382.

\bibitem{wang2016ldc}
\BIBentryALTinterwordspacing
Z.~Wang, H.~Mi, and A.~Ittycheriah, ``Sentence similarity learning by lexical
  decomposition and composition,'' in \emph{Proceedings of COLING 2016, the
  26th International Conference on Computational Linguistics: Technical
  Papers}.\hskip 1em plus 0.5em minus 0.4em\relax The COLING 2016 Organizing
  Committee, 2016, pp. 1340--1349. [Online]. Available:
  \url{http://aclweb.org/anthology/C16-1127}
\BIBentrySTDinterwordspacing

\bibitem{mikolov2013-distributed}
\BIBentryALTinterwordspacing
T.~Mikolov, I.~Sutskever, K.~Chen, G.~S. Corrado, and J.~Dean, ``Distributed
  representations of words and phrases and their compositionality,'' in
  \emph{Advances in Neural Information Processing Systems 26}, C.~J.~C. Burges,
  L.~Bottou, M.~Welling, Z.~Ghahramani, and K.~Q. Weinberger, Eds.\hskip 1em
  plus 0.5em minus 0.4em\relax Curran Associates, Inc., 2013, pp. 3111--3119.
  [Online]. Available: \url{http://arxiv.org/abs/1310.4546}
\BIBentrySTDinterwordspacing

\bibitem{mikolov2013-efficient}
\BIBentryALTinterwordspacing
T.~Mikolov, K.~Chen, G.~Corrado, and J.~Dean, ``Efficient estimation of word
  representations in vector space,'' \emph{CoRR}, vol. abs/1301.3781, 2013.
  [Online]. Available: \url{http://arxiv.org/abs/1301.3781}
\BIBentrySTDinterwordspacing

\bibitem{rodriguesetal2016}
J.~Rodrigues, B.~António, N.~Steven, and S.~João, ``{Word Embeddings
  Resources for the Portuguese Language},'' in \emph{Computational Processing
  of the Portuguese Language: 12th International Conference (PROPOR-2016)},
  J.~Silva, R.~Ribeiro, P.~Quaresma, A.~Adami, and A.~Branco, Eds.\hskip 1em
  plus 0.5em minus 0.4em\relax Springer International Publishing, 2016.

\end{thebibliography}

\end{document}